%% file: root.tex
\documentclass[letterpaper, 10pt, conference]{ieeeconf}
\usepackage{pgfplotstable} 
\usepackage{graphicx}
\usepackage{cite}       
\usepackage{xcolor}
\usepackage{amsmath,amsfonts,amssymb}
\usepackage{siunitx}
\usepackage{multirow}
\usepackage{url}
\usepackage{tabularray}
\usepackage{pgfplots}
\usetikzlibrary{patterns}
\pgfplotsset{compat=newest}
\usepgfplotslibrary{statistics}
\usepgfplotslibrary{fillbetween}
\usetikzlibrary{intersections} 
\usepgfplotslibrary{groupplots}
\usepackage{array}

\usepackage{tikz}             
\usepackage{circuitikz}
\usetikzlibrary{patterns,hobby,decorations.pathmorphing,arrows.meta,calc}

\let\labelindent\relax
\usepackage{enumitem}
\usepackage{nomencl}

\usetikzlibrary{patterns,intersections}
\usepgfplotslibrary{statistics,fillbetween,groupplots}
\usepackage{pgfplotstable}

\newcommand{\ie}{\textit{i.e.}}
\newcommand{\eg}{\textit{e.g.}}

\begin{document}
\title{Towards Estimating Normal and Shear Interface Pressures in Prosthetic Sockets via Least Squares and Mechanics Modeling}
\author{Axel Gonz\'alez Cornejo, Tianhao Yu, Chi Hwan Lee, and Edgar Bol\'ivar-Nieto}

\markboth{}{}
\maketitle
\begin{abstract}
Prosthetic socket fitting remains largely manual and iterative, and objective fit metrics are still limited. Part of the challenge is the lack of long-term real-life pressure data at the residual limb--socket interface. Traditional pressure sensors are prone to drift over time, and capture only normal pressures at sparse locations within the socket, missing a critical component for biomechanical analysis: shear. Although some sensors can report both normal and shear interface stresses, these components are often difficult to decouple because of measurement crosstalk. One potential path forward is to develop models that can augment available measurements. This work introduces a testbed to evaluate model performance under sparse pressure sensing using two complementary validation signals: (i) the global wrench (\ie, total forces and moments expressed in an orthonormal frame) transmitted through the socket, by an artificial residual-limb, and (ii) local interface loads (\ie, decoupled normal and shear pressure components in a right-hand-rule orthogonal frame that lives in each instrumented location) measured by sparse sensing clusters, each composed of four capacitance-sensing channels. Rather than presenting full-field pressure estimates, the focus is on an analysis sequence that quantifies how well candidate mechanical models explain both global and local measurements under controlled conditions. A quasi-static spring--mass contact model is evaluated, and its parameters are identified via a two-stage convex least-squares problem. Validation under static loading shows that estimating constant bias terms reduces steady offsets in the wrench channels and improves agreement with local measurements. A Pareto-front sensitivity analysis further illustrates how the trade-off between global and local objectives changes when bias terms are included.
\end{abstract}

\section{Introduction}
\label{sec:intro}

The prosthetic socket is the primary mechanical interface between a lower-limb prosthesis and the user. Since the earliest biomechanical analyses of socket loading \cite{radcliffe1955functional, cw1962biomechanics}, it has been recognized that when loads are not distributed appropriately at the residual limb--socket interface, discomfort and skin/tissue complications can occur, and severe cases may require clinical intervention \cite{neumann_concepts_2005, portnoy_patient-specific_2009, nerone_reamputation_2013, rathnayake_lower_2020}. Early in-socket pressure studies further demonstrated that both the magnitude and distribution of interface stresses vary across socket designs and activity conditions \cite{appoldt_stump-socket_1968, naeff_-relationship_nodate, beil2002interface}. Understanding interface loading is therefore important for improving socket comfort and safety. Despite decades of clinical practice and research, socket fabrication and fitting remain largely manual and iterative. Common workflows still rely on casting, hand rectification, and repeated adjustments guided by prosthetist judgment and user feedback \cite{moerman_automated_2016}, which makes outcomes difficult to standardize and document across clinics. A variety of socket design philosophies have been explored for different amputation levels \cite{brodie_transfemoral_2022}, yet objective criteria for what constitutes a ``good'' fit remain ambiguous; existing fit metrics are often subjective in nature \cite{hanspal_prosthetic_2003, larsen_performance_2021}, and efforts toward systematic socket adjustment are still work in progress \cite{weathersby_automatic_2021}. These gaps make it difficult to evaluate new fabrication methods, materials, and fitting tools using objective evidence \cite{mak_state---art_2001, olsen_socket_documentation_2022}.

Another part of the challenge is obtaining accurate and complete measurements of residual limb--socket interface stresses. Although in-socket sensing has been widely explored using diverse transduction principles (\eg, capacitive, piezoelectric, piezoresistive) \cite{ko_scoping_2021, el-sayed_piezoelectric_2014, lee_conductive_2015, tabor_textile-based_2021}, three persistent limitations constrain practical use. First, sensor drift over extended use degrades reliability and repeatability \cite{polliack_scientific_2000, buis_calibration_problems_1997}. Second, most instrumented studies have focused on normal pressures, even though shear stresses are clinically relevant and have been associated with deep tissue injury---a precursor to pressure ulcers and tissue necrosis---since multi-axis measurements are typically more difficult to obtain and validate than normal pressure alone \cite{zhang_frictional_1996, laszczak_pressure_2016, kwak_wireless_2020}. Third, sensor placement is limited by both hardware characteristics and geometry: the curved, irregular, and compliant limb--socket interface often restricts where sensors can be installed, resulting in spatially sparse measurements that incompletely represent the pressure distribution. Moreover, noncompliant sensing hardware can perturb the local stress field, leading to boundary effects and normal--shear crosstalk (\ie, an apparent shear response under a purely normal load, and vice versa) \cite{polliack_scientific_2000}. Recent advances in flexible sensor arrays have expanded the design space for wearable interface monitoring \cite{fan_machine-knitted_2020, lee_highly_2018, ghonasgi_modular_2021}, and purpose-built interfacial stress sensors have been validated for prosthetic environments \cite{laszczak_development_2015, ibarra_aguila_interface_2020}. Nevertheless, studies spanning socket types and activity conditions (e.g., static standing, level walking, stair/slope ambulation) continue to show that interface loads can vary substantially with socket design and gait phase \cite{goh_static_2003, dou_pressure_2006, wolf_pressure_2009}, reinforcing the need for rigorous sensor validation and repeatable measurement protocols when using sparse in-socket measurements \cite{polliack_scientific_2000}.

Physics-based models are a natural complement to sparse sensing, but their use depends on credible experimental validation. Prior reviews have emphasized both the potential and the challenges of socket interface modeling, including uncertainties in contact conditions and soft-tissue behavior \cite{mak_state---art_2001, silver-thorn_review_nodate}. Related work in wearable-device biomechanics has shown that calibrated contact models can predict interaction forces accurately when suitable experimental data are available \cite{serrancoli_subject-exoskeleton_2019}, and combined kinematic--kinetic analyses at the residual-limb/socket interface have been used to relate global joint loads to local interface stresses \cite{tang_combined_2017}. For model development to be applicable, experiments must provide synchronized ground-truth signals that test a model at multiple levels, including global load transmission and local interface behavior. The main contribution of this work is a testbed to evaluate model performance under sparse pressure sensing using two complementary validation signals: (i) the global wrench (\ie, total forces and moments expressed in a right-hand-rule orthonormal frame) transmitted through the socket, by an artificial residual-limb, and measured by a ground-referenced 6-axis load cell mounted below the socket, and (ii) local interface loads measured by embedded sensing clusters---$2\times2$ groups of custom-built textile sensors that have demonstrated robustness in prosthetic applications \cite{yu_embroidered_sensors_under_review}---reported as decoupled normal and shear pressure components in a right-hand-rule orthonormal frame rigidly attached to each cluster location.
 
The remainder of this paper is organized as follows. Section~\ref{sec:testbed} describes the testbed design, including the socket and pylon assemblies, the pressure-sensing clusters, the load cell, and the motion-capture-based synchronization scheme. Section~\ref{sec:modeling} introduces the quasi-static lumped-parameter contact model and its offset variant. Section~\ref{sec:optimization} formulates the two-stage convex parameter-identification procedure, including the pre-optimization of rest-configuration parameters and the multiobjective stiffness estimation with physically meaningful constraints. Section~\ref{sec:results} presents validation results under static loading, comparing stiffness-only and stiffness-plus-bias estimators at both the global wrench and local force levels, together with a Pareto-front sensitivity analysis. Finally, section~\ref{sec:discussion} discusses the findings and outlines future work directions.

\section{Testbed Design}\label{sec:testbed}

\begin{figure}[t]
    \centering
    \includegraphics[width=0.7\linewidth]{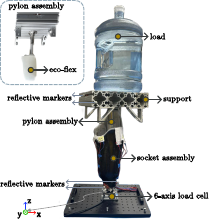}
    \caption{Testbed for controlled socket--limb contact, allowing synchronous global wrench and local interface monitoring.}
    \label{fig:testbed}
\end{figure}

\begin{figure}
    \centering
    \includegraphics[width=0.7\linewidth]{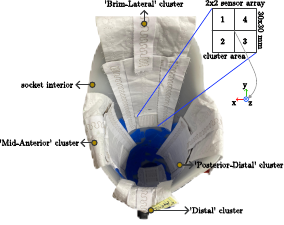}
    \caption{Pressure-sensing on the socket interior (Brim--Lateral, Mid--Anterior, Posterior--Distal, and Distal). Each cluster is a \(2\times2\) capacitive array (\SI{30}{\milli\meter}\(\times\)\SI{30}{\milli\meter}) and reports stresses in a right-hand-rule local frame fixed to the cluster. The sensor fabrication and testing in prosthetic applications constitutes prior work \cite{yu_embroidered_sensors_under_review}.}
    \label{fig:clusters}
\end{figure}

Validation of the mechanical models of the residual-limb--socket interface (Section~\ref{sec:results}) requires synchronized ground-truth measurements at two levels: a global wrench (overall force and moment balance) and local interface loads (spatial load distribution). To obtain both under repeatable conditions, a testbed was built with two assemblies: a \emph{Socket assembly} and a \emph{Pylon assembly} (Fig.~\ref{fig:testbed}). During experiments, the Pylon assembly’s artificial limb is inserted into the prosthetic socket to create a controlled contact condition that mimics donning. The Socket assembly includes a 6-axis load cell (M3564F, Sunrise Instruments Co., Ltd., Shanghai, China) mounted below the socket to provide the global wrench reference, and four pressure-sensing clusters mounted on the socket interior to provide local load references. Each cluster contains four capacitive sensors arranged in a $2\times2$ grid (Fig.~\ref{fig:clusters}). The Pylon assembly includes a load support, a pylon, and an artificial residual limb cast from a positive mold of an adult-size transfemoral socket. The limb is made of Eco-flex silicone rubber (Smooth-On, Inc., Macungie, PA, USA) to provide skin--like compliant deformation under compression, and a Lycra-based ply sock is placed over the limb to reduce friction during don/doff, similar to clinical practice.

\section{Mechanical modeling of residual limb--socket interface}\label{sec:modeling}

\begin{figure}
  \centering
  \resizebox{0.9\linewidth}{!}{\input{contact_model_schematic.tex}}
  \caption{Schematic of the lumped-parameter contact model (shown in 2D for visualization). A lumped-mass at the current position $\mathbf{q}$ is connected to socket wall anchors $\{\mathbf{a}_i\}$ through unilateral springs with stiffnesses $\{\mathbf{k}_i\}$ and rest position $\{\mathbf{q}_{r}\}$ (\ie, the position where the spring model generates forces consistent with load cell and cluster sensor readings in the absence of an external load). The full model in this work is three-dimensional: each spring is tri-axial, and the anchor locations, stiffnesses, and rest positions are vectors \(\in \mathbb{R}^3\).}
  \label{fig:toyz}
\end{figure}
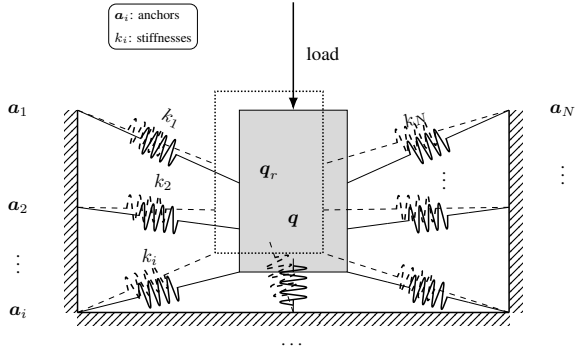

\subsection{Points of interest, anchors, and coordinate frames}

The residual limb--socket contact is represented by $n$ discrete \emph{points of interest} on the socket interior. Each point of interest $i\in\{1,\dots,n\}$ is associated with an \emph{anchor point} $\mathbf{a}_i\in\mathbb{R}^3$ fixed on the socket wall (Fig.~\ref{fig:toyz}). The set $\{\mathbf{a}_i\}$ was generated by discretizing a 3D scan of the socket interior in CAD (Fusion~360) and tiling the surface with the maximum number of non-overlapping $30\times 30$~mm squares, matching the footprint of each pressure-sensing cluster so that modeled and measured quantities are defined at comparable spatial locations. The anchor $\mathbf{a}_i$ is taken as the center of tile $i$.

All modeled wrenches are expressed in the \emph{socket/load-cell frame} $\{S\}$, whose origin is at the load-cell origin and whose axes follow the load-cell right-hand-rule convention. Each sensing cluster (or, more generally, each point of interest) has its own \emph{local frame} $\{L_i\}$ rigidly attached to the tile/cluster, in which the reconstructed local forces are reported.

\paragraph{Rotation between $\{L_i\}$ and $\{S\}$.}
Let ${}^{S}\mathbf{R}_{L,i}\in SO(3)$ map a vector expressed in $\{L_i\}$ into $\{S\}$:
\begin{equation}
{}^{S}\mathbf{v} = {}^{S}\mathbf{R}_{L,i}\,{}^{L_i}\mathbf{v},
\qquad
{}^{L_i}\mathbf{v} = ({}^{S}\mathbf{R}_{L,i})^\top\,{}^{S}\mathbf{v}.
\label{eq:frame_map}
\end{equation}
In this work, ${}^{S}\mathbf{R}_{L,i}$ is constructed from unit axes of the local frame expressed in $\{S\}$:
\begin{equation}
{}^{S}\mathbf{R}_{L,i} =
\begin{bmatrix}
{}^{S}\hat{\mathbf{x}}_i & {}^{S}\hat{\mathbf{y}}_i & {}^{S}\hat{\mathbf{z}}_i
\end{bmatrix}.
\label{eq:R_from_axes}
\end{equation}
These unit vectors can be obtained directly from CAD for each tile: ${}^{S}\hat{\mathbf{z}}_i$ is chosen as the inward surface normal of tile $i$, and ${}^{S}\hat{\mathbf{x}}_i$ is chosen as a tangent direction (e.g., aligned with one tile edge or a consistent ``proximal'' CAD direction). The third axis is enforced by the right-hand rule,
\begin{equation}
{}^{S}\hat{\mathbf{y}}_i =
\frac{{}^{S}\hat{\mathbf{z}}_i \times {}^{S}\hat{\mathbf{x}}_i}
{\left\|{}^{S}\hat{\mathbf{z}}_i \times {}^{S}\hat{\mathbf{x}}_i\right\|}.
\label{eq:y_from_cross}
\end{equation}
With this construction, ${}^{S}\mathbf{R}_{L,i}$ is orthonormal and right-handed by design.

\subsection{Kinematics and rest configuration}

Socket--limb interaction is modeled as a single lumped mass connected to the socket wall through a set of piecewise linear springs. The lumped mass represents the residual bone motion, while each spring represents compliant soft tissue that can transmit compressive loads to the socket wall when the limb is donned and loaded. Springs are \emph{unilateral}: compression produces force, while tension produces no force.

At discrete-time sample $k$, let $\mathbf{q}[k]\in\mathbb{R}^3$ denote the lumped-mass position expressed in $\{S\}$. Each spring $i$ has a rest-configuration parameter $\mathbf{q}_{r,i}\in\mathbb{R}^3$ (also expressed in $\{S\}$). This vector defines the position at which spring $i$ is at zero load along each axis of $\{S\}$, and it captures local clearance/preload effects that vary across the socket interior. The relative displacement is
\begin{equation}
\boldsymbol{\delta}_i[k] = \mathbf{q}[k] - \mathbf{q}_{r,i}.
\label{eq:delta_def_reorg}
\end{equation}

\subsection{Unilateral tri-axial spring law}

To enforce unilateral contact smoothly and axis-wise, an elementwise contact gate is defined as
\begin{equation}
\mathbf{g}(\boldsymbol{\delta}_i)=\tfrac{1}{2}\mathbf{1}-\tfrac{1}{2}\tanh(\alpha \boldsymbol{\delta}_i),
\label{eq:contact_gate_reorg}
\end{equation}
where $\alpha>0$ controls the sharpness of the transition, $\tanh(\cdot)$ is applied elementwise, and $\mathbf{1}\in\mathbb{R}^3$ is a vector of ones. For $\delta_{i,j}<0$ (compression along axis $j$), $g(\delta_{i,j})\rightarrow 1$; for $\delta_{i,j}>0$ (tension), $g(\delta_{i,j})\rightarrow 0$.

The spring force at point of interest $i$ is defined elementwise as
\begin{equation}
{}^{S}\mathbf{F}_i[k]
=
\mathbf{g}(\boldsymbol{\delta}_i[k])\odot \Big(\mathbf{k}_i\odot \boldsymbol{\delta}_i[k]\Big),
\label{eq:model1_reorg}
\end{equation}
where ${}^{S}\mathbf{F}_i[k]\in\mathbb{R}^3$ is the modeled contact force applied to the socket at anchor $\mathbf{a}_i$, $\mathbf{k}_i=[k_{x,i}\;k_{y,i}\;k_{z,i}]^\top$ contains axis-wise stiffness parameters, and $\odot$ denotes the elementwise product.

\subsection{Predicted global wrench and local forces}

Under quasi-static conditions, the modeled global wrench at the load cell is obtained from force and moment balance:
\begin{equation}
{}^{S}\mathbf{w}_{\mathrm{model}}[k]
=
\begin{bmatrix}
{}^{S}\mathbf{M}_{\mathrm{model}}[k]\\
{}^{S}\mathbf{F}_{\mathrm{model}}[k]
\end{bmatrix}
=
\begin{bmatrix}
\sum_{i=1}^{n} \mathbf{a}_i \times {}^{S}\mathbf{F}_i[k]\\[2pt]
\sum_{i=1}^{n} {}^{S}\mathbf{F}_i[k]
\end{bmatrix},
\label{eq:wrench_model_reorg}
\end{equation}
where ${}^{S}\mathbf{w}_{\mathrm{model}}[k]\in\mathbb{R}^6$ and $\times$ denotes the cross product. Local forces are collected by stacking
\begin{equation}
{}^{S}\mathbf{f}_{\mathrm{model}}[k]
=
\begin{bmatrix}
({}^{S}\mathbf{F}_1[k])^\top & \cdots & ({}^{S}\mathbf{F}_n[k])^\top
\end{bmatrix}^\top
\in\mathbb{R}^{3n}.
\label{eq:local_force_stack_reorg}
\end{equation}

Because reconstructed cluster forces are reported in $\{L_i\}$, comparisons are performed in a common frame using \eqref{eq:frame_map}. For example, measured local forces can be rotated into the socket/load-cell frame via
\begin{equation}
{}^{S}\mathbf{F}_{\mathrm{meas},i}[k] = {}^{S}\mathbf{R}_{L,i}\,{}^{L_i}\mathbf{F}_{\mathrm{meas},i}[k],
\label{eq:meas_to_socket}
\end{equation}
or, equivalently, modeled forces can be rotated into $\{L_i\}$ via ${}^{L_i}\mathbf{F}_{\mathrm{model},i}[k]=({}^{S}\mathbf{R}_{L,i})^\top {}^{S}\mathbf{F}_i[k]$.

\subsection{Model variant with offset}

A second variant augments \eqref{eq:model1_reorg} with an axis-wise offset:
\begin{equation}
{}^{S}\mathbf{F}_i[k]
=
\mathbf{g}(\boldsymbol{\delta}_i[k])\odot \Big(\mathbf{k}_i\odot \boldsymbol{\delta}_i[k]\Big)+ \mathbf{b}_i,
\label{eq:model2_reorg}
\end{equation}
where $\mathbf{b}_i\in\mathbb{R}^3$ is an axis-wise constant offset. Under a linear spring law, this term can be interpreted as a preload or as an equivalent shift in the rest position:
\begin{equation}
\mathbf{k}_i\odot\boldsymbol{\delta}_i + \mathbf{b}_i
=
\mathbf{k}_i\odot(\boldsymbol{\delta}_i-\boldsymbol{\delta}_{0,i}),
\qquad
\boldsymbol{\delta}_{0,i}=-\mathbf{b}_i\oslash\mathbf{k}_i,
\label{eq:offset_equiv_reorg}
\end{equation}
where $\oslash$ denotes elementwise division. Accordingly, $\mathbf{b}_i$ and $\mathbf{q}_{r,i}$ are not independent without additional constraints or regularization.

\section{Convex Optimization}\label{sec:optimization}

Parameters were identified in two stages via weighted least squares. First, a \emph{pre-optimization} step estimated the springs' rest-position vector~$\mathbf{q}_{r,\mathrm{vec}}$. Second, with the rest positions fixed, stiffness parameters (and, optionally, bias terms) were estimated. Both stages share the same objective structure: a weighted sum of wrench-fit, local-force-fit, and spatial-smoothness terms.

\subsection{Common notation}

Let $N$ denote the number of discrete-time samples. The per-sample wrench ${}^{S}\mathbf{w}_{\mathrm{model}}[k]\in\mathbb{R}^{6}$ and local-force vector ${}^{S}\mathbf{f}_{\mathrm{model}}[k]\in\mathbb{R}^{3n}$ from \eqref{eq:wrench_model_reorg}--\eqref{eq:local_force_stack_reorg} are stacked over $k=1,\dots,N$ to form the following column vectors: 
\begin{equation}
\begin{aligned}
\mathbf{w}_{\mathrm{meas}}
&:=
\begin{bmatrix}
\mathbf{w}_{\mathrm{meas}}[1]^\top & \cdots & \mathbf{w}_{\mathrm{meas}}[N]^\top
\end{bmatrix}^\top
\in \mathbb{R}^{6N},\\
\mathbf{f}_{\mathrm{meas}}
&:=
\begin{bmatrix}
\mathbf{f}_{\mathrm{meas}}[1]^\top & \cdots & \mathbf{f}_{\mathrm{meas}}[N]^\top
\end{bmatrix}^\top
\in \mathbb{R}^{3nN},
\end{aligned}
\label{eq:stacked_meas}
\end{equation}
and the corresponding model predictions are stacked identically into $\mathbf{w}_{\mathrm{model}}$ and $\mathbf{f}_{\mathrm{model}}$.

\subsection{Stage~1: Rest-configuration estimation}\label{sec:stage1}

The decision variable is $\mathbf{q}_{r,\mathrm{vec}}\in\mathbb{R}^{p}$, which collects all rest-configuration parameters into a single vector (any parameter sharing used to reduce $p$ is omitted for clarity).

Stiffness is fixed to a pre-characterized value $k_0$ and bias terms are disabled. Under these conditions the spring-force expression~\eqref{eq:model1_reorg} simplifies to
\begin{equation}
\mathbf{F}_i[k] = k_0\big(\mathbf{q}[k]-\mathbf{q}_{r,i}\big)
= k_0\,\mathbf{q}[k] \;-\; k_0\,\mathbf{q}_{r,i},
\label{eq:Fi_factorization}
\end{equation}
which is affine in the unknown $\mathbf{q}_{r,i}$. Stacking over all springs and samples, and substituting into~\eqref{eq:wrench_model_reorg}--\eqref{eq:local_force_stack_reorg}, yields
\begin{equation}
\begin{aligned}
\mathbf{w}_{\mathrm{model}}(\mathbf{q}_{r,\mathrm{vec}})
&= \mathbf{w}_0 \;-\; \mathbf{A}_{\mathrm{w}}\,\mathbf{q}_{r,\mathrm{vec}},
\\
\mathbf{f}_{\mathrm{model}}(\mathbf{q}_{r,\mathrm{vec}})
&= \mathbf{f}_0 \;-\; \mathbf{A}_{\mathrm{f}}\,\mathbf{q}_{r,\mathrm{vec}},
\end{aligned}
\label{eq:model_affine_qr}
\end{equation}
where $\mathbf{w}_0$ and $\mathbf{f}_0$ depend only on known quantities ($\mathbf{q}[k]$, anchors $\mathbf{a}_i$, rotations ${}^{S}\mathbf{R}_{L,i}$), and the matrices $\mathbf{A}_{\mathrm{w}}\in\mathbb{R}^{6N\times p}$ and $\mathbf{A}_{\mathrm{f}}\in\mathbb{R}^{3nN\times p}$ collect the coefficients multiplying $\mathbf{q}_{r,\mathrm{vec}}$. No assumption is introduced beyond regrouping terms from~\eqref{eq:Fi_factorization}.

Moving the constant terms to the measurement side,
\begin{equation}
\tilde{\mathbf{w}} := \mathbf{w}_{\mathrm{meas}}-\mathbf{w}_0,
\qquad
\tilde{\mathbf{f}} := \mathbf{f}_{\mathrm{meas}}-\mathbf{f}_0,
\label{eq:tilde_defs}
\end{equation}
both data-fit residuals take the standard form $\mathbf{A}\mathbf{q}_{r,\mathrm{vec}}-\tilde{\mathbf{y}}$.

Physically feasible bounds are enforced using the maximum observed compression $\Delta_{\max}\approx 0.008~\mathrm{m}$ (measured between the distal ends of pylon and Eco-flex). Each rest-position block $\mathbf{q}_{r,i}$ is constrained around a reference position $\mathbf{q}_{\mathrm{ref}}\in\mathbb{R}^3$:
\begin{equation}
\mathbf{q}_{\mathrm{ref}}-\Delta_{\max}\mathbf{1}
\;\le\;
\mathbf{q}_{r,i}
\;\le\;
\mathbf{q}_{\mathrm{ref}}+\Delta_{\max}\mathbf{1},
\qquad \forall\, i,
\label{eq:qr_box}
\end{equation}
where inequalities are elementwise.

The pre-optimization problem is then
\begin{equation}
\begin{aligned}
\min_{\mathbf{q}_{r,\mathrm{vec}}}\;&
\lambda_{\mathrm{w}}\,
\left\|\mathbf{A}_{\mathrm{w}}\mathbf{q}_{r,\mathrm{vec}}-\tilde{\mathbf{w}}\right\|_2^2
\\
&+\;
\lambda_{\mathrm{f}}\,
\left\|\mathbf{A}_{\mathrm{f}}\mathbf{q}_{r,\mathrm{vec}}-\tilde{\mathbf{f}}\right\|_2^2
\;+\;
\left\|\mathbf{D}\mathbf{q}_{r,\mathrm{vec}}\right\|_2^2
\\
\text{s.t. }\;& \eqref{eq:qr_box},
\end{aligned}
\label{eq:preopt_compact}
\end{equation}
where $\lambda_{\mathrm{w}}\ge 0$ and $\lambda_{\mathrm{f}}\ge 0$ weight the wrench-fit and local-force-fit terms, respectively, and $\mathbf{D}$ is a fixed finite-difference operator that penalizes variation between neighboring rest parameters.

\subsection{Stage~2: Stiffness estimation}\label{sec:stage2}

With the rest positions fixed to the pre-optimized value $\mathbf{q}_{r,\mathrm{vec}}^\star$, the displacements $\boldsymbol{\delta}_i[k]$ from~\eqref{eq:delta_def_reorg} and the contact gates $\mathbf{g}(\boldsymbol{\delta}_i[k])$ from~\eqref{eq:contact_gate_reorg} are known quantities. The spring-force expression~\eqref{eq:model1_reorg} is therefore linear in the stiffness vector $\mathbf{k}_{\mathrm{vec}}$, and the stacked model predictions can be written as
\begin{equation}
\begin{aligned}
\mathbf{w}_{\mathrm{model}}(\mathbf{k}_{\mathrm{vec}}) &= \mathbf{A}_{\mathrm{w}}\,\mathbf{k}_{\mathrm{vec}},
\\
\mathbf{f}_{\mathrm{model}}(\mathbf{k}_{\mathrm{vec}}) &= \mathbf{A}_{\mathrm{f}}\,\mathbf{k}_{\mathrm{vec}},
\end{aligned}
\label{eq:linear_k_model}
\end{equation}
where the matrices $\mathbf{A}_{\mathrm{w}}$ and $\mathbf{A}_{\mathrm{f}}$ now collect the known displacement-and-gate products that multiply $\mathbf{k}_{\mathrm{vec}}$ (analogous to~\eqref{eq:model_affine_qr}, obtained by the same regrouping procedure).

The stiffness estimation problem has the same weighted-least-squares structure as~\eqref{eq:preopt_compact}, with a nonnegativity constraint replacing the box constraint:
\begin{equation}
\begin{aligned}
\min_{\mathbf{k}_{\mathrm{vec}}}\;&
\lambda_{\mathrm{w}}\,
\left\|\mathbf{A}_{\mathrm{w}}\mathbf{k}_{\mathrm{vec}}-\mathbf{w}_{\mathrm{meas}}\right\|_2^2
\\
&+\;
\lambda_{\mathrm{f}}\,
\left\|\mathbf{A}_{\mathrm{f}}\mathbf{k}_{\mathrm{vec}}-\mathbf{f}_{\mathrm{meas}}\right\|_2^2
\;+\;
\left\|\mathbf{D}\mathbf{k}_{\mathrm{vec}}\right\|_2^2
\\
\text{s.t. }\;&\mathbf{k}_{\mathrm{vec}} \succeq \mathbf{0},
\end{aligned}
\label{eq:multiobj_k_compact}
\end{equation}
where $\mathbf{D}$ is a finite-difference operator (analogous to the one in~\eqref{eq:preopt_compact}, now defined over neighboring stiffness parameters) and the weights $\lambda_{\mathrm{w}},\lambda_{\mathrm{f}}\ge 0$ control the trade-off between wrench fit, local-force fit, and spatial smoothness.

\medskip
\paragraph{Stiffness-plus-bias variant.}
Following the offset model~\eqref{eq:model2_reorg}, a second variant augments the decision variables with stacked bias vectors $\mathbf{b}_{\mathrm{w}}$ and $\mathbf{b}_{\mathrm{f}}$ (consistent with $\mathbf{w}_{\mathrm{meas}}$ and $\mathbf{f}_{\mathrm{meas}}$, respectively). The model predictions become
\begin{equation}
\begin{aligned}
\mathbf{w}_{\mathrm{model}}(\mathbf{k}_{\mathrm{vec}},\mathbf{b}_{\mathrm{w}}) &= \mathbf{A}_{\mathrm{w}}\,\mathbf{k}_{\mathrm{vec}}+\mathbf{b}_{\mathrm{w}},
\\
\mathbf{f}_{\mathrm{model}}(\mathbf{k}_{\mathrm{vec}},\mathbf{b}_{\mathrm{f}}) &= \mathbf{A}_{\mathrm{f}}\,\mathbf{k}_{\mathrm{vec}}+\mathbf{b}_{\mathrm{f}},
\end{aligned}
\label{eq:linear_k_bias_model}
\end{equation}
and the optimization retains the same structure as~\eqref{eq:multiobj_k_compact} with $\mathbf{b}_{\mathrm{w}}$ and $\mathbf{b}_{\mathrm{f}}$ added to the decision variables:
\begin{equation}
\begin{aligned}
\min_{\mathbf{k}_{\mathrm{vec}},\,\mathbf{b}_{\mathrm{w}},\,\mathbf{b}_{\mathrm{f}}}\;&
\lambda_{\mathrm{w}}\,
\left\|\mathbf{A}_{\mathrm{w}}\mathbf{k}_{\mathrm{vec}}+\mathbf{b}_{\mathrm{w}}-\mathbf{w}_{\mathrm{meas}}\right\|_2^2
\\
&+\;
\lambda_{\mathrm{f}}\,
\left\|\mathbf{A}_{\mathrm{f}}\mathbf{k}_{\mathrm{vec}}+\mathbf{b}_{\mathrm{f}}-\mathbf{f}_{\mathrm{meas}}\right\|_2^2
\;+\;
\left\|\mathbf{D}\mathbf{k}_{\mathrm{vec}}\right\|_2^2
\\
\text{s.t. }\;&\mathbf{k}_{\mathrm{vec}} \succeq \mathbf{0}.
\end{aligned}
\label{eq:multiobj_k_bias_compact}
\end{equation}
The smoothness term penalizes only stiffness variation; the bias vectors are unconstrained.

\medskip
\paragraph{Convexity.}
Problems \eqref{eq:preopt_compact}, \eqref{eq:multiobj_k_compact}, and \eqref{eq:multiobj_k_bias_compact} are convex quadratic programs: each objective term is a squared $\ell_2$ norm of an affine function of the decision variables, and all constraints are linear~\cite{boyd_convex_2023}.
\section{Experimental Protocol}\label{sec:experimental}
During testing, the Pylon assembly (\SI{9.3}{kg}) was inserted into the Socket assembly and a fixed load of \SI{19}{kg} was applied through the pylon's support to create a nearly steady socket--limb contact condition. To track the pose of both assemblies relative to a world coordinate frame, three reflective markers were attached to each rigid body. Their trajectories were recorded using an OptiTrack motion-capture system (NaturalPoint, Inc., Corvallis, OR, USA), following an approach similar to marker-based methods used for measuring residual-limb/socket motion \cite{childers_marker-based_2016}. Each trial consisted of holding the load for approximately \SI{10}{min} and then removing it; small vibrations in marker-tracking data and load cell readings were observed immediately after load placement and removal. The reflective-marker trajectories were used in post-processing to reconstruct rigid-body coordinate frames, which then allowed us to express points of interest (e.g., sensor locations) in local coordinates. For validation, a representative segment ($N=1000$ frames) with static load maintained was extracted from the static dataset. The segment was defined as the longest contiguous interval that (i) contains no intended loading changes, (ii) exhibits a near-constant mean wrench (steady plateau), and (iii) preserves valid samples across most channels (limited missing data). Selection was based on these steadiness and data-quality criteria and not on model fit.

\subsection{Hardware Communication Protocols}\label{sec:commprotocol}
All devices in the measurement chain---I2C bus (pressure clusters), SPI bus (load cell), and Wi-Fi protocol (Motive motion-capture software, NaturalPoint, Inc., Corvallis, OR, USA)---were synchronized using a common trigger generated by the microcontroller unit, a Raspberry Pi 5 board (Raspberry Pi Ltd., Cambridge, UK). This trigger started the load cell acquisition, the pressure-cluster acquisition, and motion-capture recording together. Data were logged at \SI{4}{Hz} for the experiments reported in this paper.

\subsection{Going from capacitance recordings to normal and shear pressure data}
Normal pressure is obtained from capacitance using our calibration: for each cluster, the mean capacitance across its four sensors is computed and mapped to normal pressure using a first-degree polynomial fit. Shear pressure is computed from the relative differences within the $2\times2$ grid (see Fig.~\ref{fig:clusters}). First, the cluster mean is subtracted from each sensor signal. Then, opposing sensor pairs are compared (left vs.\ right, and top vs.\ bottom) to obtain two in-plane shear components. These components are converted to shear pressure using a total-degree-2 surface fit. Both the normal- and shear-pressure conversions follow the sensor designer’s recommended calibration procedure \cite{yu_embroidered_sensors_under_review}. 
\section{Results}\label{sec:results}

Two estimator variants are compared: (i) tri-axial stiffness values only, with offset terms fixed to zero \eqref{eq:multiobj_k_compact}, and (ii) stiffness-plus-bias, with constant bias terms estimated jointly with stiffnesses \eqref{eq:multiobj_k_bias_compact}. In all plots, measured signals are overlaid with model predictions, and per-component errors are reported as RMSE and MAE over all frames.

\subsection{Global Wrench Agreement}

\begin{figure}[t]
    \centering
    \includegraphics[width=\linewidth]{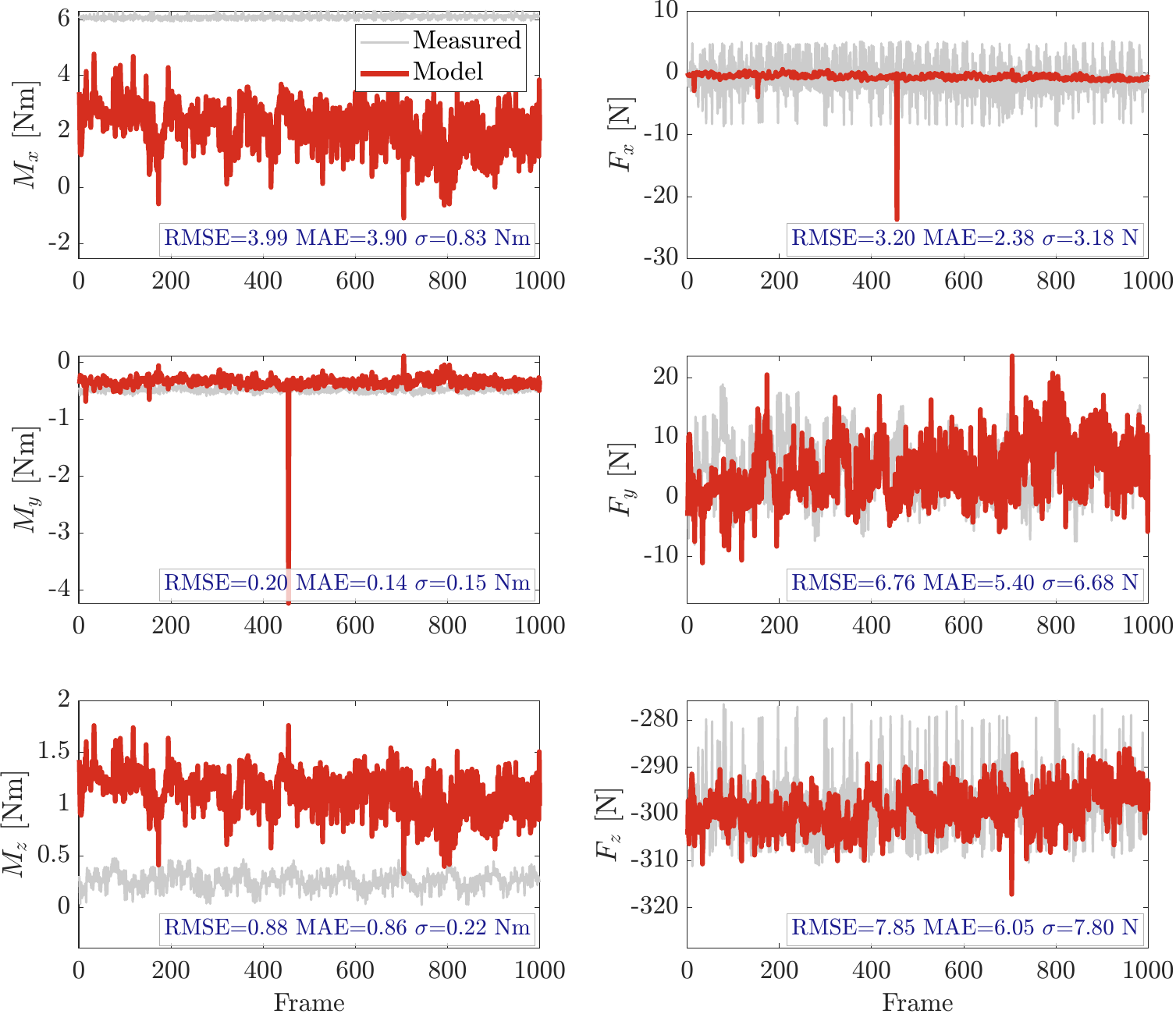}
    \caption{Global wrench: measured (gray) vs.\ model (red) for the stiffness-only estimator (bias $= 0$, $N=1000$ frames). Per-component RMSE, MAE, and residual $\sigma$ are shown in blue.}
    \label{fig:wrench_agreement0b}
\end{figure}

\begin{figure}[t]
    \centering
    \includegraphics[width=\linewidth]{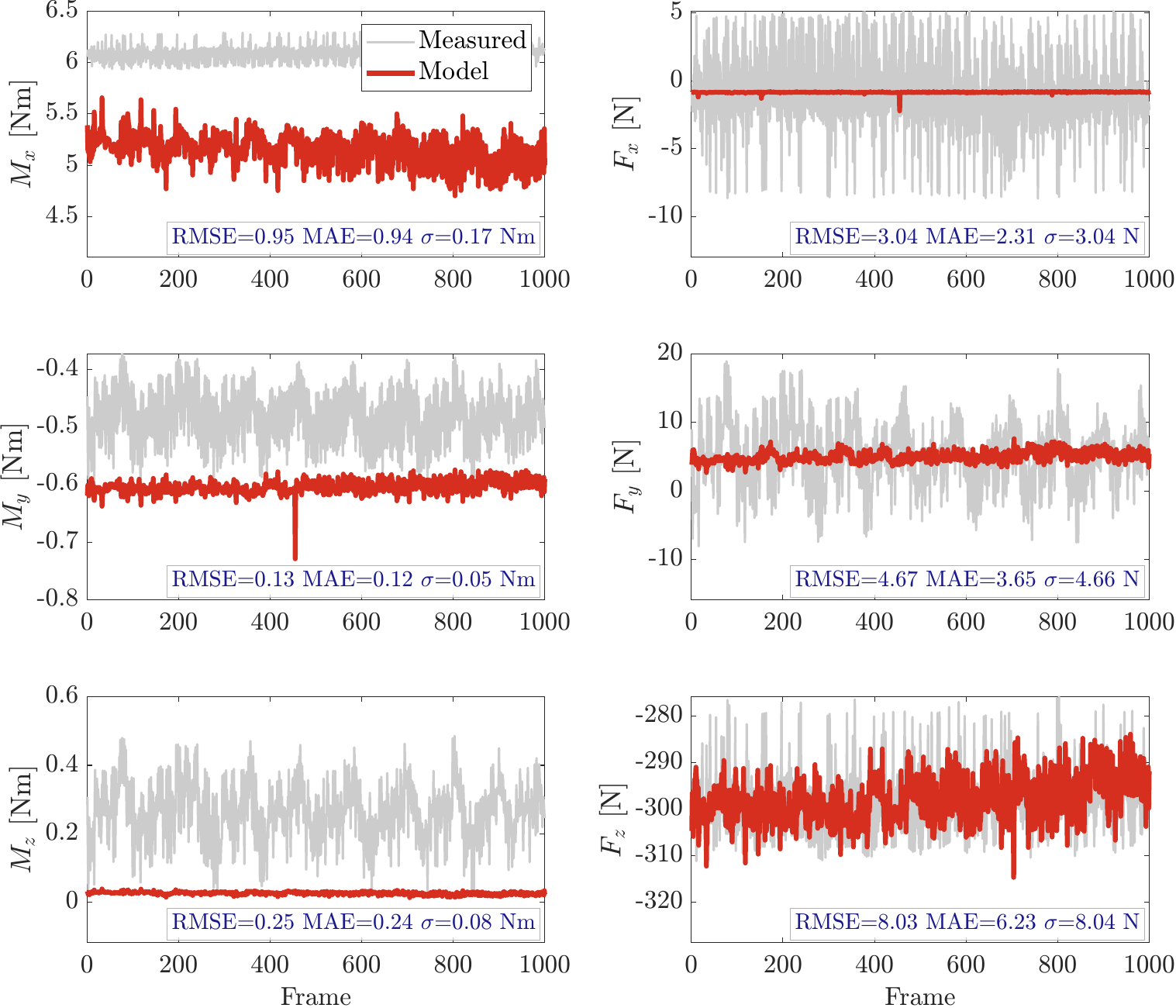}
    \caption{Global wrench: measured vs.\ model for the stiffness-plus-bias estimator over the same trial. Plot conventions match Fig.~\ref{fig:wrench_agreement0b}.}
    \label{fig:wrench_agreement}
\end{figure}

Global feasibility was evaluated by comparing the measured load-cell wrench with the model prediction (Figs.~\ref{fig:wrench_agreement0b} and~\ref{fig:wrench_agreement}). Without bias (Fig.~\ref{fig:wrench_agreement0b}), errors are dominated by $M_x$ (RMSE~$=$~\SI{3.99}{Nm}) and by $F_z$ and $F_y$ (RMSE~$=$~\SI{7.85}{\newton} and \SI{6.76}{\newton}). The mean force and moment RMSE are \SI{5.94}{\newton} and \SI{1.69}{Nm}. Because $\sigma \ll \text{RMSE}$ (e.g., $\sigma=$~\SI{0.83}{Nm} vs.\ RMSE~$=$~\SI{3.99}{Nm} for $M_x$), most of the mismatch appears as a steady offset.

With bias terms (Fig.~\ref{fig:wrench_agreement}), moment RMSE falls below \SI{1}{Nm} for all components ($M_x$: \SI{0.95}{Nm}, $M_y$: \SI{0.13}{Nm}, $M_z$: \SI{0.25}{Nm}), reducing the mean moment RMSE to \SI{0.44}{Nm}. The mean force RMSE decreases modestly to \SI{5.25}{\newton}, with the remaining mismatch dominated by $F_z$ (RMSE~$=$~\SI{8.03}{\newton}). Here $\sigma\approx\text{RMSE}$, indicating that the residual is primarily time-varying rather than a constant offset.
\subsection{Local Measurements Agreement}

\begin{figure*}[t]
  \centering
  \includegraphics[width=0.9\textwidth]{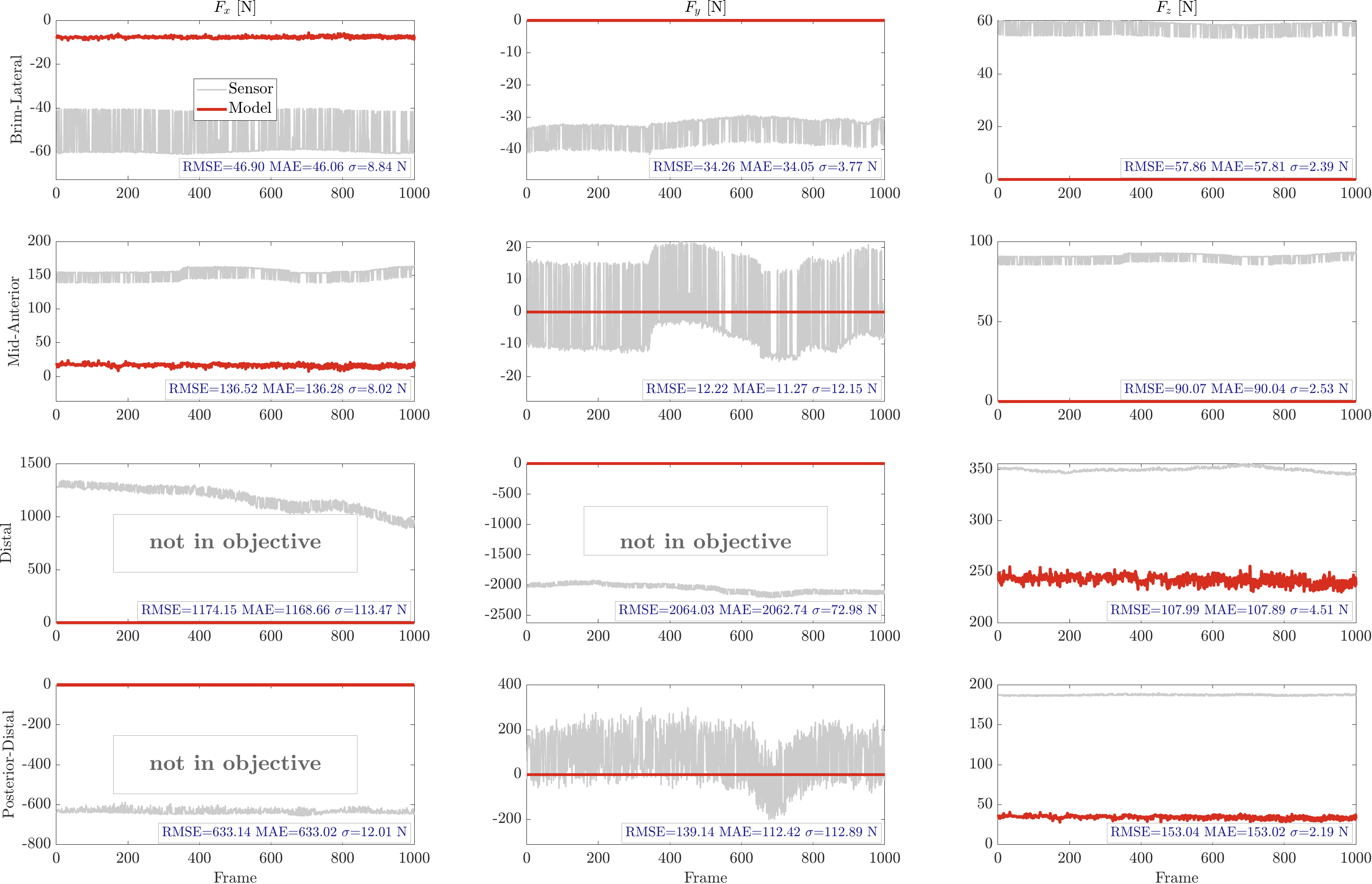}
  \caption{Local forces at the four pressure-sensor clusters: measured (gray) vs.\ model (red) for the stiffness-only estimator (bias $=0$). Channels marked ``not in objective'' were excluded from the cost function. Per-channel RMSE, MAE, and $\sigma$ are reported.}
  \label{fig:cluster_agreement0b}
\end{figure*}

\begin{figure*}[t]
  \centering
  \includegraphics[width=0.9\textwidth]{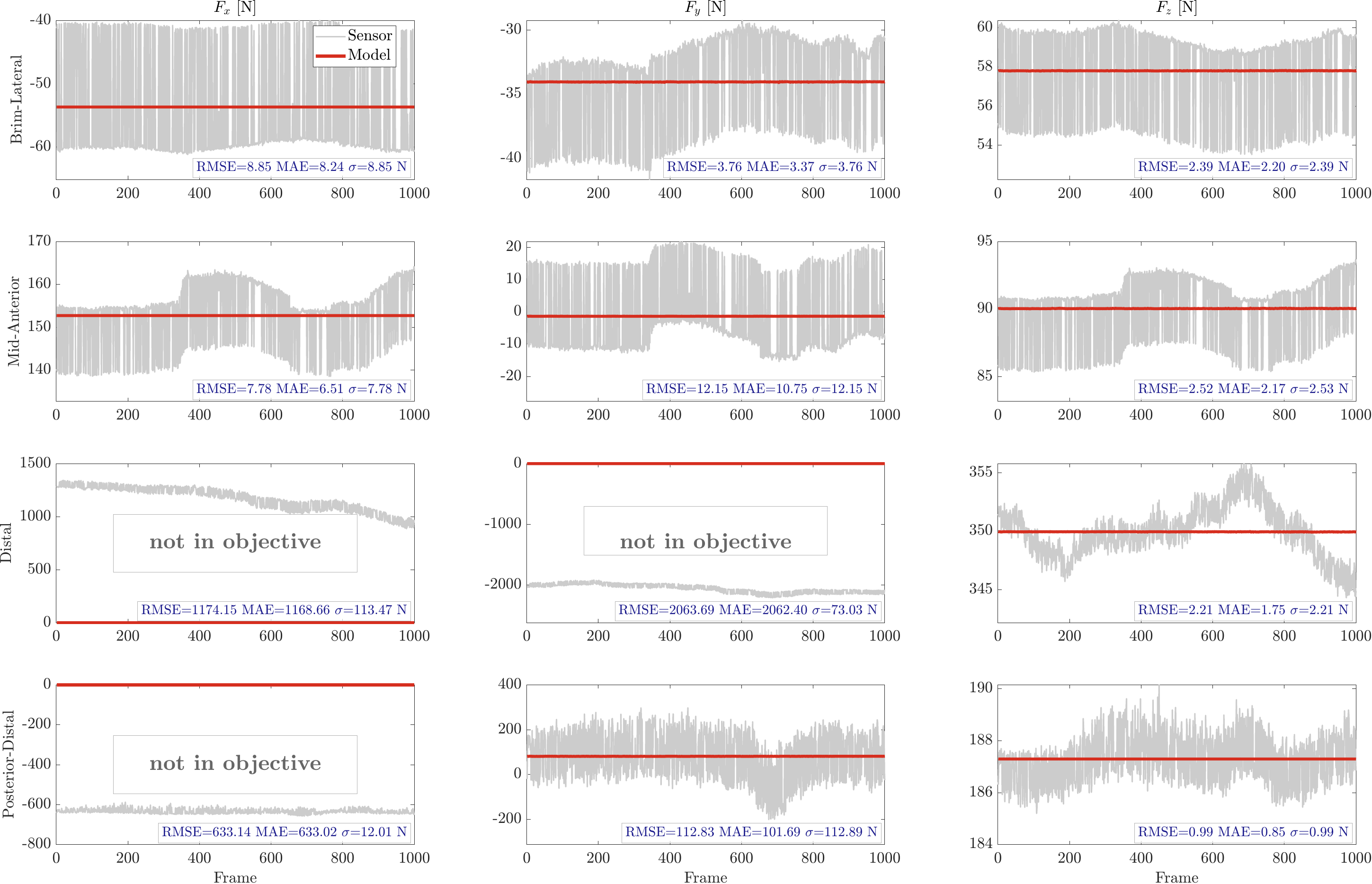}
  \caption{Local forces for the stiffness-plus-bias estimator over the same trial. Plot conventions match Fig.~\ref{fig:cluster_agreement0b}.}
  \label{fig:cluster_agreement}
\end{figure*}

Local agreement is evaluated by comparing reconstructed cluster forces with model-predicted forces at the same locations (Figs.~\ref{fig:cluster_agreement0b} and~\ref{fig:cluster_agreement}). Three shear channels (distal $F_x$, distal $F_y$, and posterior-distal $F_x$) were excluded from the objective because their reconstructed magnitudes were implausible (e.g., distal shear peaks above \SI{1000}{\newton} under an applied load of about \SI{300}{\newton}). These channels are labeled ``not in objective'' in the figures.

For the remaining channels in the bias-free fit (Fig.~\ref{fig:cluster_agreement0b}), large constant offsets appear in both normal and shear components. Normal $F_z$ RMSE is \SI{57.86}{\newton}, \SI{90.07}{\newton}, \SI{107.99}{\newton}, and \SI{153.04}{\newton} across clusters; representative shear errors include brim-lateral $F_x$ and $F_y$ RMSE of \SI{46.90}{\newton} and \SI{34.26}{\newton}, and mid-anterior $F_x$ RMSE of \SI{136.52}{\newton}. Consistent with the global wrench, $\sigma$ is small relative to RMSE, supporting an offset-dominated error.

With bias (Fig.~\ref{fig:cluster_agreement}), most normal components drop to the few-newton range ($F_z$ RMSE of \SI{2.39}{\newton}, \SI{2.52}{\newton}, \SI{2.21}{\newton}, and \SI{0.99}{\newton}). Most shear channels also improve (e.g., brim-lateral $F_x$ RMSE \SI{8.85}{\newton}; mid-anterior $F_x$ RMSE \SI{7.78}{\newton}), although posterior-distal $F_y$ remains high (RMSE~$=$~\SI{112.83}{\newton}). Over all objective-included channels, the median RMSE decreases from \SI{90.07}{\newton} (bias $=0$) to \SI{3.76}{\newton} (with bias).
\subsection{Sensitivity Analysis}

\begin{figure}[t]
    \centering
    \includegraphics[width=0.95\linewidth]{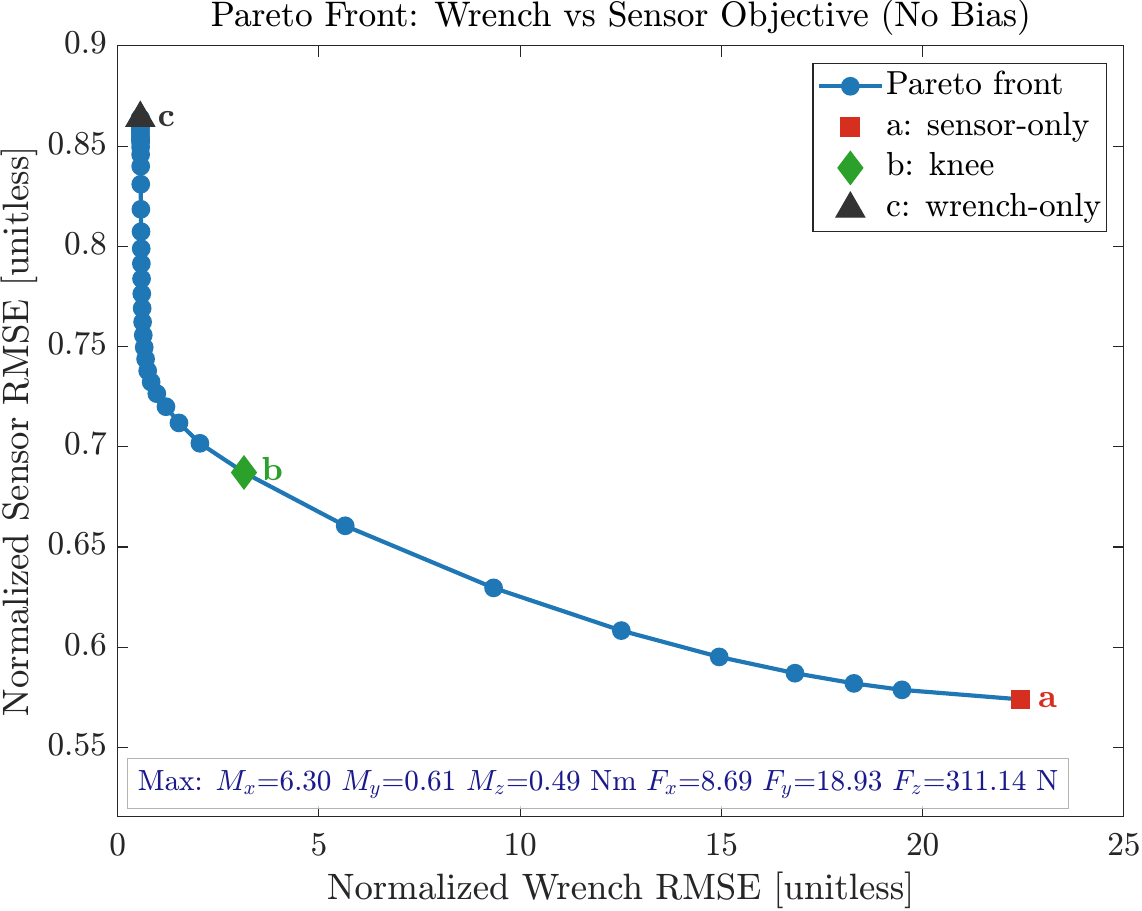}
    \caption{Pareto front (normalized wrench RMSE vs.\ normalized sensor RMSE) for the bias-free estimator. Markers \textbf{a}, \textbf{b}, \textbf{c} denote the sensor-only, knee, and wrench-only solutions, respectively.}
    \label{fig:pareto_front0b}
\end{figure}

\begin{figure}[t]
    \centering
    \includegraphics[width=0.95\linewidth]{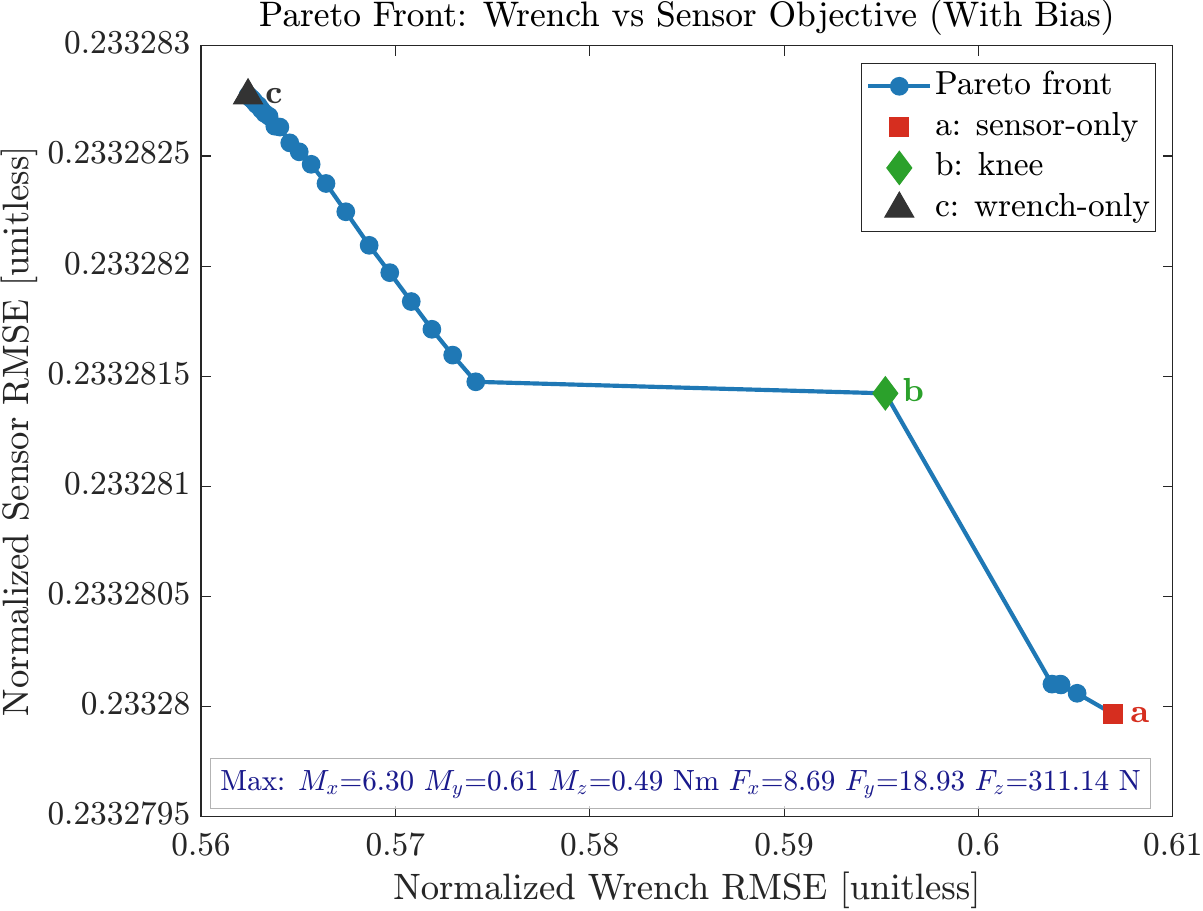}
    \caption{Pareto front for the stiffness-plus-bias estimator. Axes are zoomed relative to Fig.~\ref{fig:pareto_front0b}; the tight non-dominated set indicates a weak trade-off~\cite[p.~182]{boyd_convex_2023}. Marker conventions match Fig.~\ref{fig:pareto_front0b}.}
    \label{fig:pareto_front_bias}
\end{figure}

A Pareto-front analysis assessed sensitivity to the weighting $\theta\in[0,1]$ between wrench and sensor objectives. Post-fit errors are reported as \emph{normalized} RMSE to avoid unit mismatch between moment and force magnitudes.

In the bias-free case (Fig.~\ref{fig:pareto_front0b}), a pronounced knee separates a region where both objectives improve together from one where improving either requires disproportionate degradation of the other, confirming that $\theta$ materially affects the solution. With bias (Fig.~\ref{fig:pareto_front_bias}), the non-dominated set collapses to a tight neighborhood (note zoomed axes), indicating that the two objectives are nearly compatible for this static segment.

\section{Discussion and Conclusion}\label{sec:discussion}

\subsection{Discussion}

Estimating constant bias terms removes steady offsets that the stiffness-only model cannot capture. The largest gains are in moments and normal components: for example, $M_x$ RMSE decreases from \SI{3.99}{Nm} to \SI{0.95}{Nm}, and brim-lateral $F_z$ RMSE decreases from \SI{57.86}{\newton} to \SI{2.39}{\newton}. This suggests preload-like effects or fixed offsets due to calibration baselines, frame-registration errors, or unmodeled constant loads. Similar offset issues have been reported in residual-limb/socket measurement systems, especially when shear is included~\cite{laszczak_pressure_2016,wheeler_pressure_2016,sanders_residual_2011}.

The local results also show that fit quality depends on the upstream estimate of the rest-position vector $\mathbf{q}_{r,\mathrm{vec}}$. Because spring engagement depends on compression, small pose errors change which springs ``get activated'' to contribute to the global and local optimization objectives, limiting what stiffness-only estimation can achieve. Bias terms improve agreement but do not correct the underlying engagement pattern, so recovered stiffnesses may be less directly interpretable.

The Pareto fronts are consistent: a clear knee in the bias-free case indicates a real trade-off between objectives, while the tight non-dominated set with bias shows that both objectives can be satisfied simultaneously once constant offsets are permitted. In practice, $\theta$ selection matters in the bias-free formulation but is inconsequential with bias for static plateaus.

\subsection{Conclusion and Future Work}
A testbed for validating models of residual-limb--socket interaction was presented. On a static load 'hold' segment, estimating constant bias terms reduced the median local-force RMSE from \SI{90.07}{\newton} to \SI{3.76}{\newton} and the mean moment RMSE from \SI{1.69}{Nm} to \SI{0.44}{Nm}. The remaining accuracy is limited by imperfect estimation of the rest positions $\mathbf{q}_{r,\mathrm{vec}}$, which affects spring engagement.

Future work will focus on: (i) improving the equilibrium-pose estimate by refining the pre-optimization stage; and (ii) reconstructing pressures/forces at non-instrumented locations toward a full-socket pressure map.
\section*{Acknowledgment}
 Research reported in this publication was supported by the National Institute Of Biomedical Imaging And Bioengineering of the National Institutes of Health under Award Number R21EB034879. The content is solely the responsibility of the authors and does not necessarily represent the official views of the National Institutes of Health. 
 
\bibliographystyle{IEEEtran}
\bibliography{references}

\end{document}

%% file: contact_model_schematic.tex
\begin{circuitikz}

\pattern[pattern=north east lines] (-0.25,2) rectangle (0,-1.75);
\draw[thick] (0,2) -- (0,-1.75);

\pattern[pattern=north east lines] (0,-2) rectangle (8,-1.75);
\draw[thick] (0,-1.75) -- (8,-1.75);

\pattern[pattern=north east lines] (8.25,2) rectangle (8,-1.75);
\draw[thick] (8,2) -- (8,-1.75);

\coordinate (mSW) at (3,-1);
\coordinate (mNE) at (5,2);

\draw[fill=gray!30] (mSW) rectangle (mNE);
\node at ($(mSW)!0.5!(mNE) + (0,-0.55)$) {$\boldsymbol{q}$};

\def\dx{-0.45}
\def\dy{0.35}

\coordinate (mSWr) at ($(mSW)+(\dx,\dy)$);
\coordinate (mNEr) at ($(mNE)+(\dx,\dy)$);

\draw[thick, densely dotted] (mSWr) rectangle (mNEr);
\node at ($(mSWr)!0.5!(mNEr)$) {$\boldsymbol{q}_r$};

\coordinate (Ltop) at (3,0.65);
\coordinate (Lmid) at (3,-0.2);
\coordinate (Lbot) at (3,-1);

\coordinate (Cbot) at (4,-0.75);

\coordinate (Rbot) at (5,-1);
\coordinate (Rmid) at (5,-0.2);
\coordinate (Rtop) at (5,0.65);

\coordinate (LtopR) at ($(Ltop)+(\dx,\dy)$);
\coordinate (LmidR) at ($(Lmid)+(\dx,\dy)$);
\coordinate (LbotR) at ($(Lbot)+(\dx,\dy)$);

\coordinate (CbotR) at ($(Cbot)+(\dx,\dy)$);

\coordinate (RbotR) at ($(Rbot)+(\dx,\dy)$);
\coordinate (RmidR) at ($(Rmid)+(\dx,\dy)$);
\coordinate (RtopR) at ($(Rtop)+(\dx,\dy)$);

\coordinate (A1)  at (0,2);
\coordinate (A2)  at (0,0.2);
\coordinate (AiL) at (0,-1.75);

\coordinate (AiB) at (4,-1.75);

\coordinate (AiR) at (8,-1.75);
\coordinate (AjR) at (8,0.2);
\coordinate (AN)  at (8,2);

\node at (-1.1,2) {$\boldsymbol{a}_1$};
\node at (-1.1,0.2) {$\boldsymbol{a}_2$};
\node at (-1.1,-0.75) {$\vdots$};
\node at (-1.1,-1.75) {$\boldsymbol{a}_i$};
\node at (9.0,0.9) {$\vdots$};
\node at (9.0,2) {$\boldsymbol{a}_N$};

\draw (A1) -- (0.75, 1.6625);
\draw (0.75, 1.6625) to[spring, l=$k_{1}$] (2.25,0.9875);
\draw (2.25,0.9875) -- (Ltop);

\draw (A2) -- (0.75,0.1);
\draw (0.75,0.1) to[spring, l=$k_{2}$] (2.25,-0.1);
\draw (2.25,-0.1) -- (Lmid);


\draw (AiL) -- (0.75, -1.5625);
\draw (0.75, -1.5625) to[spring, l=$k_{i}$] (2.25, -1.1875);
\draw (2.25, -1.1875) -- (Lbot);

\node at (6.8,0.80) {$\vdots$};

\draw (AiB) -- (4,-1.5);
\draw (4,-1.5) to[spring] (4,-1);
\draw (4,-1) -- (Cbot);

\draw (Rbot) -- (5.75, -1.1875);
\draw (5.75, -1.1875) to[spring] (7.25, -1.5625);
\draw (7.25, -1.5625) -- (AiR);

\draw (Rmid) -- (5.75, -0.1);
\draw (5.75, -0.1) to[spring] (7.25, 0.1);
\draw (7.25, 0.1) -- (AjR);

\draw (Rtop) -- (5.75, 0.9875);
\draw (5.75, 0.9875) to[spring, l=$k_{N}$] (7.25, 1.6625);
\draw (7.25, 1.6625) -- (AN);

\draw[dashed] (A1)  to[spring] (LtopR);
\draw[dashed] (A2)  to[spring] (LmidR);
\draw[dashed] (AiL) to[spring] (LbotR);

\draw[dashed] (AiB) to[spring] (CbotR);

\draw[dashed] (AiR) to[spring] (RbotR);
\draw[dashed] (AjR) to[spring] (RmidR);
\draw[dashed] (AN)  to[spring] (RtopR);

\draw[thick, -{Latex}] (4,4) -- (4,2);
\node at (4.55,3.05) {load};

\node at (4,-2.35) {$\cdots$};

\node[draw, rounded corners, fill=white, align=left, inner sep=3pt]
at (1.4,3.55) {\scriptsize $\boldsymbol{a}_i$: anchors\\ \scriptsize $k_i$: stiffnesses};

\end{circuitikz}